\documentclass[conference]{IEEEtran}
\IEEEoverridecommandlockouts

\usepackage{cite}
\usepackage{amsmath,amssymb,amsfonts}
\usepackage{algorithmic}
\usepackage{graphicx}
\usepackage{textcomp}
\usepackage{comment}

\usepackage{subcaption}

\usepackage{multirow}

\usepackage{colortbl}
\usepackage[table]{xcolor}
\usepackage{booktabs}

\definecolor{lightgreen}{RGB}{192, 242, 145}
\definecolor{lightyellow}{RGB}{255, 255, 153}
\definecolor{lightpurple}{RGB}{220, 210, 235}
\definecolor{lightorange}{RGB}{255, 204, 153}

\renewcommand\IEEEkeywordsname{Keywords}

\usepackage{tikz}
\usepackage{import}
\usetikzlibrary{calc}
\usetikzlibrary{pgfplots.groupplots}
\usepackage{pgfplots}
\usepackage{pgfplotstable}
\usepackage{graphics}
\usepgfplotslibrary{fillbetween}
\usetikzlibrary{patterns}
\usepackage[export]{adjustbox}
\usepackage{pgfplotstable}

\definecolor{darkgray176}{RGB}{176,176,176}
\definecolor{darkturquoise0191191}{RGB}{0,191,191}
\definecolor{green01270}{RGB}{0,127,0}
\definecolor{lavender204204255}{RGB}{204,204,255}
\definecolor{lemonchiffon255255204}{RGB}{255,255,204}
\definecolor{lightgoldenrodyellow229255204}{RGB}{229,255,204}
\definecolor{lightgray204}{RGB}{204,204,204}

\def\BibTeX{{\rm B\kern-.05em{\sc i\kern-.025em b}\kern-.08em
    T\kern-.1667em\lower.7ex\hbox{E}\kern-.125emX}}
\begin{document}

\title{Extending Affordance Segmentation for Wearable Robots to Multi-Object Images}

\author{\IEEEauthorblockN{Simone Lugani, Matteo Montani, Paolo Gastaldo, Rodolfo Zunino, Edoardo Ragusa}
\IEEEauthorblockA{DITEN, University of Genoa, Genoa, Italy; contact: simone.lugani@edu.unige.it}}

\IEEEpubid{%
  \parbox[t]{\columnwidth}{%
    \footnotesize
    \raggedright
    \copyright~2026 IEEE.
    Personal use of this material is permitted.
    Permission from IEEE must be obtained for all other uses, in any current
    or future media, including reprinting/republishing this material for
    advertising or promotional purposes, creating new collective works, for
    resale or redistribution to servers or lists, or reuse of any copyrighted
    component of this work in other works.\\[1pt]
    \emph{Published in IEEE COINS 2026.}
    DOI: 10.1109/COINS69995.2026.11715600.
  }%
  \hspace{\columnsep}%
  \makebox[\columnwidth]{}%
}
\maketitle

\begin{abstract}
This paper proposes a refinement to existing vision-based portable pipelines for the recognition of graspable object parts. Such pipelines can effectively support automatic grasp selection and preshaping in upper-limb prosthetic devices. However, realistic scenarios in which multiple objects are present still represent a challenge. The proposed method exploits a minor modification to the conventional hardware setup and a custom training approach to extend the applicability of vision-based pipelines. A laser pointer is introduced in the setup as an effective mechanism for object disambiguation. Accordingly, a suitable retraining of the deep convolutional neural networks supporting recognition of graspable parts is implemented. A mixed reality strategy allows the reuse of available high-quality segmentation masks to generate realistic images procedurally, extending the range of scenarios supported by the final device. Experimental results confirm that the proposed pipeline improves target disambiguation accuracy by up to 9\% compared with existing solutions, while preserving the computational cost.
\end{abstract}


\section{Introduction}
\label{sec:introduction}
Upper-limb prosthetics aim to restore motor function and improve the quality of life after amputation, but their effective control remains a challenge. Conventional myoelectric control is often sequential and cognitively demanding, especially for multi-degree-of-freedom prostheses, which limits usability and leads to high rejection rates \cite{salminger2022current}. Shared-autonomy paradigms, where the prosthesis exploits environmental sensing to assist the user during task execution \cite{human_in_the_loop_review}, alleviate these issues.

Vision-based approaches support automatic grasp selection and preshaping\cite{krausz2019survey, mouchoux2021artificial}, but existing systems rely on the user’s ability to frame the target object\cite{castro2022continuous}. This becomes problematic in realistic scenarios with multiple objects, leading to ambiguity in target selection.

We address this limitation by introducing a simple mechanism for object disambiguation in vision-based prosthetic control: a laser pointer integrated into the wearable setup. The pointer, activated via EMG signals, provides an explicit indication of the user’s intended target. This information is exploited by a lightweight affordance segmentation (AS) module implemented with Deep Neural Networks (DNNs), which identifies the object of interest and extracts its graspable regions, enabling reliable operation even in multi-object environments \cite{ragusa2021hardware,ragusa2023affordance}.


The proposed approach preserves the efficiency required for embedded deployment, while improving the robustness of target disambiguation in semi-autonomous prosthesis control. Experimental results demonstrate that the pointer enables accurate target selection and enhances system performance in challenging multi-object scenarios.

\section{Background}\label{sec:background}

The literature presents a wide range of prototypes and approaches based on teleceptive sensing devices \cite{hundhausen2019resource, starke2022semi}, including RGB cameras \cite{weiner2022designing}, radars \cite{mastinu2024explorations}, and depth sensors \cite{castro2022continuous}. Data from these sensors are integrated into autonomous control pipelines to enhance user performance \cite{castro2022continuous,rana2024affordance}. Modern robotic arms leverage hardware accelerators to enable on-board data processing \cite{hundhausen2019resource, starke2022semi,chiariotti2024future}, opening new opportunities for semi-autonomous control  \cite{tang2022wearable,sensinger2020review}.

Affordance segmentation (AS) describes the possible interactions with different parts of an object and offers valuable inputs for control pipelines \cite{jiang2021synergies, khalifa2022towards,wang2024sgsin}. Several approaches to AS have been proposed in the literature \cite{hassanin2021visual, jiang2021synergies, corona2020ganhand, khalifa2022towards,ozccil2024affordance}. However, hardware constraints and the involvement of a user in the control loop of wearable robots call for dedicated solutions. The method introduced in \cite{ragusa2021hardware} proposed a simplified and portable version of basic AS, designed to benefit from human interaction. Subsequent works further improved this approach by integrating DNN modules \cite{apicella2021affordance, lugani2023lightweight} and additional sensory inputs \cite{ragusa2023affordanceAP}. The research in \cite{ragusa2023affordance} demonstrated that such networks can run in real-time on high-end MCUs while supporting semi-autonomous control pipelines.

The integration of depth sensors \cite{krausz2019survey, chiariotti2024future} into DNN pipelines has been extensively explored \cite{deng20213d, gao2024dense, ragusa2023affordanceAP}, but it increases both computational cost and system complexity, due to modality fusion and calibration requirements \cite{zhou2024dgpinet}. 

Few works have addressed target-object identification in prosthetic shared-control pipelines. They infer the target by combining vision with user-related cues, such as prosthesis motion, depth-based reaching, or gaze and eye-hand coordination \cite{zhong2020reliable,castro2022continuous,cognolato2022improving}. Although effective, they are not formulated for AS pipelines and often rely on implicit cues, accurate framing, additional sensors, or demanding perception modules. Thus, target disambiguation remains open for portable RGB-based AS pipelines, where the visual stream should support both target selection and graspable-part segmentation.

Hardware considerations distinguish AS
implementation strategies into unconstrained, hardware-efficient, and tiny models \cite{ragusa2025filling}. These groups
loosely map to hardware platforms with different computational
capabilities. While most existing research focuses on unconstrained or tiny models, mid-range hardware-efficient solutions remain relatively unexplored. To the best of the authors’ knowledge, the works in \cite{ragusa2025filling, lugani2023lightweight, ragusa2021hardware} represent the only AS architectures specifically designed for embedded accelerators. These approaches assume a clear framing of a single object while, in this work, this limitation is addressed by introducing a direct target selection mechanism based on a laser pointer.

\section{Proposal}\label{sec:proposal}

\begin{figure}
    \centering
    \includegraphics[width=0.8\linewidth]{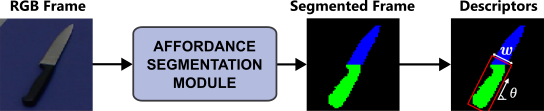}
    \caption{AS pipeline: the image from a hand-mounted camera is segmented to identify graspable parts; high-level descriptors are then extracted and used for intelligent prosthesis control.}
    \label{fig:schema_AS}
\end{figure}

\subsection{Sensing system}

In the proposed setup, AS assigns functional labels to specific object regions to identify the parts that support a given action (e.g., grasping). As shown in Figure \ref{fig:schema_AS}, the AS module processes the RGB frame acquired from the hand-held camera to generate a segmented mask of the graspable parts, e.g., a handle of an object. It enables the system to focus on that part when estimating object properties and extracting high-level descriptors, which in turn leads to a proper configuration of the hand. Thus, the AS provides automatic segmentation into relevant object parts, without explicit user instruction.

The system is designed for a semi-autonomous control pipeline, where a user relies on EMG signals for volitional control of the robotic hand. In typical semi-autonomous procedures, the user points the hand toward the object and activates, with a gesture, a trigger that initiates automatic grasping or preshaping. After the trigger, a camera captures images of the scene, and AS processes the frames to identify the graspable parts. Then, the resulting AS mask can be used to extract geometric cues (Figure \ref{fig:schema_AS}), such as an oriented bounding box on the graspable region. In some setups, a DNN automatically controls the hand actuators. In other setups, a suitable control logic exploits the output provided by the network for hand preshaping. In both cases, the DNN supporting AS should first identify the target object. 

\begin{figure}
    \centering
    \includegraphics[width=0.85\linewidth]{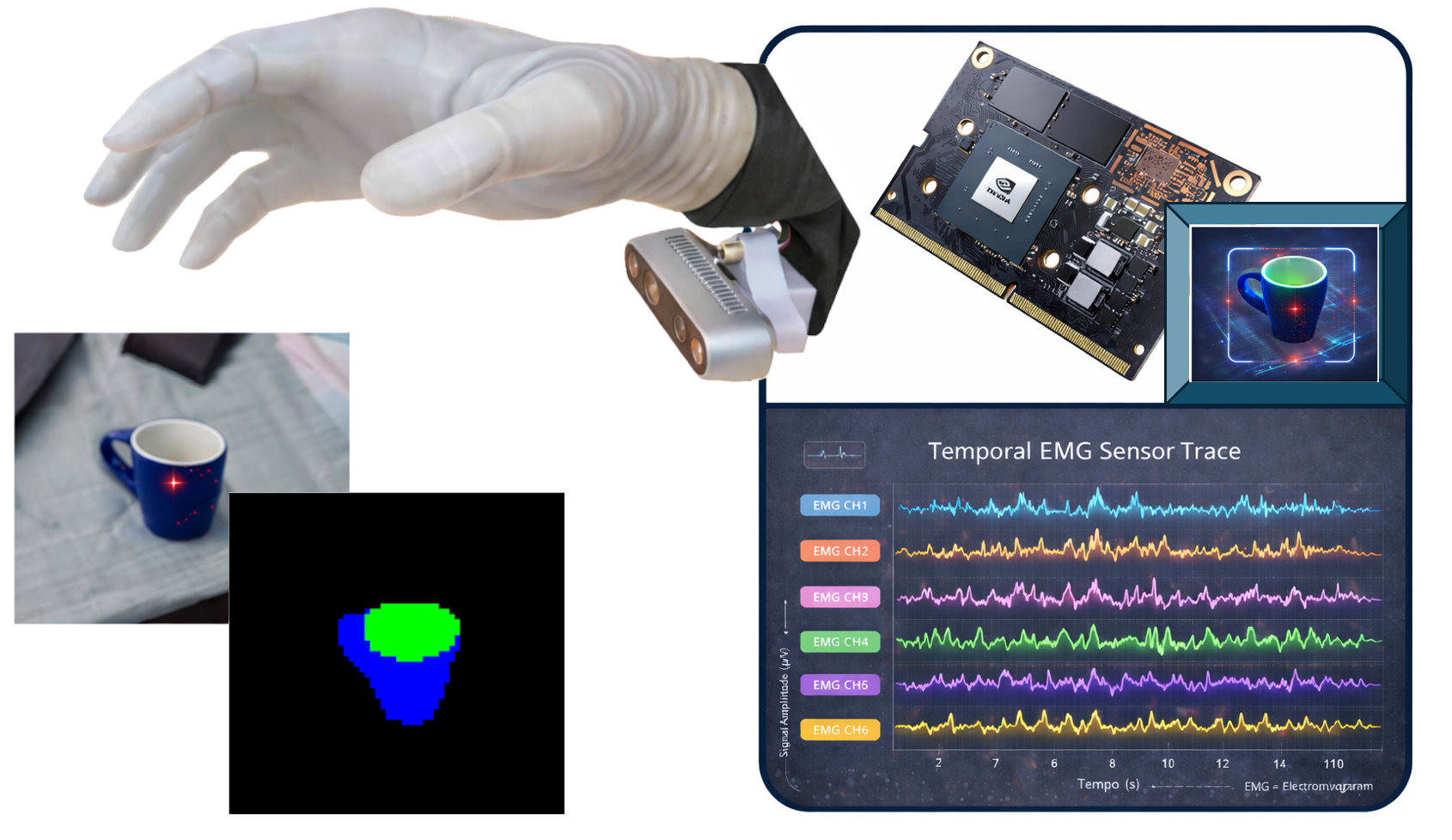}
    \caption{Proposed framework: a pointer attached near the camera adds a light spot on the image }
    \label{fig:prot}
\end{figure}

In the proposed approach, when the first trigger is sent, the system activates a laser pointer mounted on the hand and waits for a few seconds to allow the user to tune the framing, i.e., centering the pointer on the target object. Figure \ref{fig:prot} shows one possible implementation of the proposed setup, where both the RGB camera and the laser pointer are mounted beneath the hand. The AS module uses the pointer as a reference and segments the object targeted by the user. This setup extends the system to scenarios like those shown in Figure \ref{fig:main}, which are currently too challenging for state-of-the-art portable setups \cite{ragusa2025filling}. In addition, the pointer makes otherwise ambiguous target objects explicit.

\subsection{Training the AS module with a mixed reality approach}
In the proposed pipeline, the AS module should be trained to focus only on the target object, namely the one pointed by the user. Hence, one cannot rely only on standard datasets for AS that do not include pointers on the images. In addition, most of the existing datasets for AS have been generated with cameras mounted on robots leading to different viewpoints. These limitations motivate the need for a mixed-reality approach, where suitable images are augmented to generate a more representative dataset. A real pointer produces a clear light spot surrounded by smooth intensity gradients. Though full realism would require additional precautions, simple synthetic patterns can approximate its appearance.

\begin{figure}[t]
    \centering
    \includegraphics[width=0.9\linewidth]{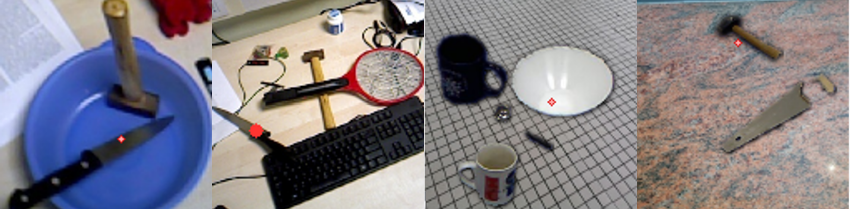}
    \caption{Examples: IIT real scenes with synthetic laser annotations (I, II), and UMD mixed-reality scenes (III, IV).}
    \label{fig:main}
\end{figure}

The most widely used dataset for AS is UMD. It contains a set of everyday objects framed from many angles. Each object is recorded under different orientations against a fixed blue background, and a corresponding segmentation mask is provided for each view. The labeling procedure produced good segmentation masks, making this dataset a standard benchmark; however, the fixed background and illumination limit the generalization capabilities of models trained on it. Using the object segmentation mask, we isolate it from the background with good precision and extract a set of object images from the original dataset. These images are further refined through post-processing based on the fixed background color, smoothing object borders. The segmented objects are then placed onto random realistic backgrounds to generate mixed-reality scenes, as shown in Fig.~\ref{fig:main}.

Another common dataset for AS is IIT-Aff (IIT). It provides rectangular bounding boxes and semantic segmentation for each object, and includes many multi-object scenes, as in the first and second images in Figure~\ref{fig:main}.

\section{Experiments}\label{sec:exp}

\begin{table}[t]
\centering
\begin{tabular}{lcccc}
\toprule
 & \multicolumn{4}{c}{\textbf{MODELS}} \\
\cmidrule(lr){2-5}
\textbf{Testing Setup} & \cellcolor{lightgreen}\textbf{No\_P} \cite{ragusa2025filling}& \cellcolor{lightyellow}\textbf{P\_OBJ} & \cellcolor{lightpurple}\textbf{P\_BG} &\cellcolor{lightorange} \textbf{P\_GP} \\
\midrule

\cellcolor{lightgreen}
No\_Pointer & \cellcolor{lightgreen}95.69 & 95.79 & 95.57 & 91.74 \\

\cellcolor{lightyellow}
P\_Object & 95.58 & \cellcolor{lightyellow}95.77 & 95.59 & 81.30 \\

\cellcolor{lightpurple}
P\_Background & 95.55 & 95.77 &\cellcolor{lightpurple} 95.59 & 85.71 \\

\cellcolor{lightorange}
P\_Graspable & 95.61 & 95.61 & 95.45 & \cellcolor{lightorange}96.18 \\


\bottomrule
\end{tabular}

\caption{Performance of different models on foreground images for different placements of the pointer.\label{tab:sing}}

\end{table}

\subsection{Setup}
The AS module uses a lightweight DNN previously embedded in AS pipelines\cite{lugani2023lightweight}. It requires 638 MFLOPs and 3.6 M parameters, with an inference latency of 80 ms and a power consumption of 6 W on a Jetson Nano. It was trained and tested using a workstation with an NVIDIA GeForce RTX 2080 Ti with Tensorflow. All networks were trained for 100 epochs using learning rate scheduling and early stopping. Testing data were never involved in parameter or hyperparameter tuning. We used weighted cross-entropy with class weights inversely proportional to class frequencies. Geometric and color augmentation was applied throughout training.

\subsection{Single object images}

The first setup, summarized in Table~\ref{tab:sing}, evaluates the pointer effect on the generalization performance over UMD foreground images, following \cite{ragusa2025filling}. Four models were trained: No\_P without any pointer; P\_OBJ, P\_BG, and P\_GP with the pointer added to the object, the background near the object, and the object graspable part, respectively. P\_BG assesses whether the model relies on the pointer even when it does not indicate the object. Each model was tested on the same-domain test set and on the three remaining domains. Thus, as per Table \ref{tab:sing}, the No\_P model has been evaluated on a test set including only images without a pointer and on the other three test sets.  In Table \ref{tab:sing}, columns denote models and rows denote test datasets. Colored cells indicate matching training/test datasets. Each entry reports the average accuracy over the three classes.

In all cases, the models achieve high accuracy. The configuration without pointers scores an average accuracy of 95.69\%. When the pointer is placed on the object or in the background, it is ignored by the network: the accuracy remains almost constant for all test sets. When the pointer is placed in the graspable part of the object in all the training images, it plays a role in the inference phase.  When tested using the corresponding test set, it leads to a slight improvement in accuracy. In the other cases, the accuracy decreases, which confirms that the network relies on the pointer.

\subsection{Multiple object images}

The second analysis in Table \ref{tab:multi} considers images with multiple objects. The columns identify the training dataset. The rows, divided into groups using a color scheme, represent the testing set. SELF identifies that the network has been tested on the test set corresponding to the training set, i.e. data from the same dataset but that were not in the training set. The green rows remark two test sets originated with UMD foreground images with only one object. The first does not contain a pointer, the second does. The last three rows (yellow) remark three test sets with 3 objects per image. The test set named 3OB contains the pointer on the target object, 3OB-NO-P does not contain the pointer, and 3OB-RND has the pointer placed in a random position, instead of having it on the object.

Four models have been trained, one for each column. The first one was trained with a dataset generated with the proposed approach (MR). The second column (MR-IIT) indicates a dataset composed of a mixture of the MR and IIT. Similarly, the third (MR-UMD) joined MR with UMD, and the last (MR-IIT-UMD) used all three datasets. For the case of UMD, only 20\% of the dataset was included to maintain the focus of the networks on images with multiple objects. In all the training images, a pointer on the target object was included. 

\begin{table}[t]
\centering
\renewcommand{\arraystretch}{1.2}
\begin{tabular}{lcccc}
\toprule
 & \textbf{MR} & \textbf{MR-IIT} & \textbf{MR-UMD} & \textbf{MR-IIT-UMD} \\
\midrule

\rowcolor{lightpurple} SELF 
    & 92.38 & 93.36 & 91.98 & 94.29 \\

\rowcolor{lightgreen}
No\_P 
    & 87.34 & 86.50 & 95.23 & 95.52 \\

\rowcolor{lightgreen}
P\_OBJ 
    & 90.49 & 93.44 & 95.24 & 95.74 \\

\rowcolor{lightyellow}3OB 
    & 90.83 & 89.98 & 90.72 & 89.62 \\

\rowcolor{lightyellow}3OB-NO-P 
    & 83.53 & 80.61 & 83.47 & 80.12 \\

\rowcolor{lightyellow}3OB-RND 
    & 62.20 & 53.97 & 60.41 & 54.21 \\

\bottomrule
\end{tabular}

\caption{Performance on images with multiple objects\label{tab:multi}}

\end{table}

All models perform well on their respective test sets, with average accuracy above 90\% for each model. On single-object foreground images, the pointer improves performance, showing that models can exploit it. The yellow section further highlights its role: accuracy stays high with a pointer, drops by 7--9\% without it, and decreases sharply when it is randomly placed, confirming that the network focuses on it.

\subsection{Pointer shape and color}

The last analysis concerns the shape and color of the pointer. The virtual addition of the pointer can be significantly different from a real-world image. To evaluate this aspect, 4 versions of the virtual pointer have been tested after training the networks with the virtual pointer used in the previous setups. 

\begin{figure}
    \centering
    \includegraphics[width=\linewidth]{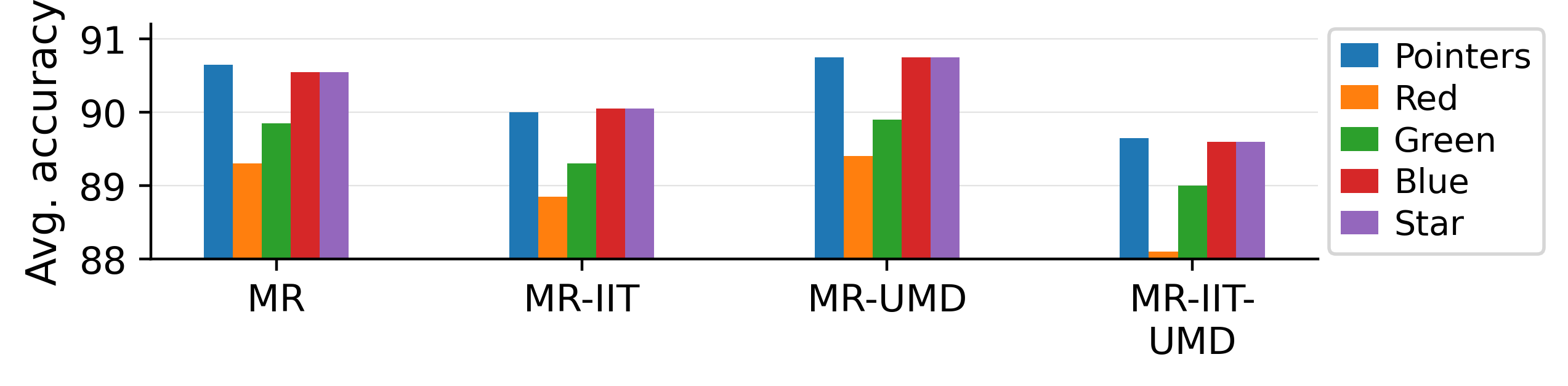}
    \caption{Effect of pointers' shape and color}
    \label{fig:pointers}
\end{figure}

Figure \ref{fig:pointers} reports the accuracy of different models for the different versions of the pointer. In the plot, Pointers denotes the virtual pointer used in the training, Red, Green, and Blue are virtual pointers with a single color. Finally, Star is a pointer roughly of the same size as the pointers used in the other images but with a specific pattern, i.e., a star. The testing set was always the set with 3 objects per image, and the pointer was placed on the target object.

The results indicate that the model is quite robust against different synthetic variations in the pointers' color and shape. In all cases, the deterioration in this challenging setup is limited: the color of the training and testing pointer is different, but still, the network can recognize the target object. This robustness to variations in pointer appearance is likely supported by the use of color augmentation during training.

\section{Conclusions}
This work introduces a laser pointer to extend portable AS pipelines for semi-autonomous prostheses to multi-object scenes. Through mixed-reality training, the system learns to identify the target object from the laser position. The results show that the model exploits the pointer to select the target object. The system proved robust to variations in pointer appearance, including changes in color and shape.

The proposal enables target disambiguation in portable vision-based prosthetic systems while preserving a lightweight pipeline, as the required change is limited to the addition of a small laser pointer. Current validation is limited to controlled and mixed-reality scenarios. In real deployments, laser visibility may be affected by ambient illumination, transparent or specular surfaces, and user-related factors. Future work should validate the approach in real operating conditions, as well as the usability and cognitive load associated with pointer-based target selection. 

\bibliographystyle{IEEEtran}
\bibliography{references}
\vfill

\end{document}